\documentclass[pdflatex,sn-apa]{sn-jnl}

\usepackage{graphicx}%
\usepackage{multirow}%
\usepackage{amsmath,amssymb,amsfonts}%
\usepackage{amsthm}%
\usepackage{mathrsfs}%
\usepackage[title]{appendix}%
\usepackage{xcolor}%
\usepackage{textcomp}%
\usepackage{manyfoot}%
\usepackage{booktabs}%
\usepackage{algorithm}%
\usepackage{algorithmicx}%
\usepackage{algpseudocode}%
\usepackage{listings}%
\usepackage{url}            
\usepackage{booktabs}       
\usepackage{amsfonts}       
\usepackage{nicefrac}       
\usepackage{svg}
\usepackage{subcaption}
\usepackage{bbm}

\theoremstyle{thmstyleone}%
\theoremstyle{thmstyletwo}%

\theoremstyle{thmstylethree}%

\newcommand{\graph}{\textsf{G}}
\newcommand{\nodeset}{\mathcal{V}}
\newcommand{\edgeset}{\mathcal{E}}
\newcommand{\motif}{\textsf{M}}
\newcommand{\mask}{{\mathbf{\hat{m}}}}
\newcommand{\gtruth}{\mathbf{m}}
\newcommand{\xaidataset}{\mathcal{D}_{\text{XAI}}}

\newcommand{\Agg}{\textsc{Agg}}
\newcommand{\Msg}{\textsc{Msg}}
\newcommand{\Upd}{\textsc{Upd}}
\newcommand{\Pool}{\textsc{Pool}}
\newcommand{\Hash}{\textsc{Hash}}
\newcommand{\NbrV}{\mathcal{N}_v}

\newcommand{\simwl}{\simeq_\textsf{WL}}
\newcommand\Multiset[1]{\left\{\!\!\!\left\{ #1 \right\}\!\!\!\right\}}

\DeclareMathOperator{\Ego}{\mathcal{S}}
\DeclareMathOperator{\freq}{\text{freq}}
\DeclareMathOperator{\argmax}{\text{argmax}}

\begin{document}

\title{A method for the systematic generation of graph XAI benchmarks via Weisfeiler--Leman coloring}


\author[1]{\fnm{Michele} \sur{Fontanesi}}\email{michele.fontanesi@phd.unipi.it}

\author[1]{\fnm{Alessio} \sur{Micheli}}\email{alessio.micheli@unipi.it}

\author[1]{\fnm{Marco} \sur{Podda}}\email{marco.podda@unipi.it}

\author*[1]{\fnm{Domenico} \sur{Tortorella}}\email{domenico.tortorella@unipi.it}

\affil[1]{\orgdiv{Department of Computer Science}, \orgname{University of Pisa}, \orgaddress{\street{Largo Bruno Pontecorvo, 3}, \city{Pisa}, \postcode{56127}, \country{Italy}}}

\abstract{
Graph neural networks (GNNs) have become the \emph{de facto} model for learning from structured data. However, the decision-making process of GNNs remains opaque to the end user, which undermines their use in safety-critical applications. Several eXplainable Artificial Intelligence (XAI) techniques for graphs have been developed to address this major issue. Focusing on graph classification, these explainers identify subgraph motifs that explain predictions. Therefore, a robust benchmarking of graph explainers is required to ensure that the produced explanations are of high quality, i.e., aligned with the GNN's decision process. However, current graph-XAI benchmarks are limited to simplistic synthetic datasets or a few real-world tasks curated by domain experts, hindering rigorous and reproducible evaluation, and consequently stalling progress in the field. To overcome these limitations, we propose a method to automate the construction of graph XAI benchmarks from generic graph classification datasets. Our approach leverages the Weisfeiler--Leman (WL) color refinement algorithm to efficiently perform approximate subgraph matching and mine class-discriminating motifs, which serve as proxy ground-truth class explanations. At the same time, we ensure that these motifs can be learned by GNNs because their discriminating power aligns with WL expressiveness. This work also introduces the \textsc{OpenGraphXAI} benchmark suite, which consists of 15 ready-made graph-XAI datasets derived by applying our method to real-world molecular classification datasets. The suite is available to the public along with a codebase to generate over 2,000 additional graph-XAI benchmarks. Finally, we present a use case that illustrates how the suite can be used to assess the effectiveness of a selection of popular graph explainers, demonstrating the critical role of a sufficiently large benchmark collection for improving the significance of experimental results.
}




\maketitle

\section{Introduction}
Graphs are data structures that allow a flexible representation of entities and their relationships as nodes and edges, respectively. Currently, the leading technology for learning from graph data is Graph Neural Networks (GNNs) \citep{BACCIU2020}. Since their inception in convolutional \citep{Micheli} or recursive \citep{Scarselli09} form, GNNs have shown promising results in a variety of applications, including drug discovery \citep{gilmer2017} and repurposing \citep{tetc2023}, protein-protein networks \citep{bmc2025}, recommender systems \citep{Wang2019}, and code analysis \citep{zhou2019}. The strength of GNNs lies in their ability of learning graph representations adaptively from data without needing feature engineering. However, as their adoption becomes widespread, so does the necessity of making their predictive mechanisms transparent to stakeholders and end-users, allowing informed decision-making in alignment with current legal and ethical standards --- such as the European Union's General Data Protection Regulation (GDPR) \citep{goodman2016eu} --- and to support the scientific discovery process \citep{XAIDrugDiscovery}.

To contribute to this objective, different eXplainable Artificial Intelligence (XAI) techniques for graph data, also known as \emph{graph explainers}, have been developed 
\citep{TaxSurvey, survey2XAI} to elucidate GNN predictions. In graph classification tasks, where GNNs learn to map entire graphs to discrete classes, an explanation corresponds to one or more subgraphs that drive the GNN prediction towards the correct class. Graph explainers try to identify these explanations by deriving real-valued importance scores for the graph components (typically nodes, but also edges or features), which can be thresholded to recover possible explanatory substructures \citep{XAI2024}. Accordingly, careful, standardized evaluation is indispensable to verify that the extracted subgraphs are aligned with the GNN's decision process through reproducible comparisons across methods, ideally on benchmarks with ground-truth motifs.

Typically, graph explainers are evaluated via the following pipeline:
\begin{enumerate}
    \item Selecting (or generating) benchmark datasets annotated with ground-truth (GT) explanations;
    \item Training a GNN on the classification task to high test accuracy;
    \item Passing the trained GNN and a test set of graphs \textit{post-hoc} to the graph explainer to obtain a set of importance scores for each graph;
    \item Quantifying the alignment between the (soft) subgraphs produced by the explainer and the annotated GTs with appropriate metrics.
\end{enumerate}
Among these steps, selecting or generating benchmark datasets with reliable GT explanations is particularly critical. Without robust benchmarks, comparative analyses become misleading, ultimately causing scientists to misdirect their research efforts. As is widely known, the abundance of high-quality benchmark datasets has arguably been a key driver of progress in fields like computer vision \citep{imagenet2009} and natural language processing \citep{glue2018}. In graph-XAI, benchmarks are typically synthetic, providing a controlled and reproducible environment for evaluating explainers. Samples in these datasets are crafted by attaching artificial GT motifs (e.g., grid graphs or house graphs, as in \cite{longa2025explaining}) to random graphs generated via preferential attachment. Although these datasets are straightforward to generate and widely adopted, they crucially lack the structural complexity and variability of real-world graphs. Conversely, only a limited number of graph-XAI datasets are derived from real-world graphs \citep{NEURIPS2020_google, varbella2024powergraph, agarwal2023evaluating}, primarily because the annotation of GT explanations demands specialized domain expertise that is often difficult to obtain. We find the current state of graph-XAI benchmarking unfit to support continuous progress. On the one hand, generated datasets are abundant but insufficiently challenging for graph explainers, leading to overestimating their true capabilities and preventing their improvement. On the other hand, the scarcity of graph-XAI datasets based on realistic graphs renders any ``field-test'' comparison among graph explainers highly sensitive to statistical fluctuations, making it hard to understand which research directions are most promising to pursue.

To close this gap and enable the systematic development of graph explainability methods, we propose a procedure to automatically construct graph-XAI benchmark datasets with integrated GT explanations derived from real-world graph classification tasks. Our approach applies the Weisfeiler--Leman (WL) coloring algorithm \citep{Weisfeiler1968} to existing graph classification datasets to identify subsets of graphs that share identical substructures detectable by GNNs within each class. We then isolate those subsets in which a given substructure acts as a discriminant for the class label, and designate that substructure as the GT explanation for the corresponding target. Notably, while the resulting benchmark dataset is essentially a synthetic restriction of the original classification task, the GT explanations are genuine structural patterns of real graphs, unlike those occurring in synthetic datasets. Furthermore, by aligning with the WL expressiveness of the message-passing GNNs \citep{Zhang2025}, we ensure that the discriminating patterns can be learned by the models \citep{Morris2019,Xu2019}. We employ this methodology to build the initial release of the \textsc{OpenGraphXAI} benchmark, a collection of 15 novel graph-XAI datasets derived from 11 established molecular graph datasets. The collection can be easily expanded to 2000+ datasets, and we provide the codebase and tools to facilitate future extensions. We demonstrate the utility of \textsc{OpenGraphXAI} by evaluating several prominent graph explainers from the literature, showing that this collection is not only useful for rigorous benchmarking but also instrumental to reasoning about the capabilities and limitations of current graph-XAI methods. We anticipate that this contribution will play a significant role in advancing progress and shaping research priorities in the field of graph-XAI, thereby enhancing the trustworthiness of graph machine learning methodologies \citep{oneto2022towards}.

This manuscript is structured as follows: in Section~\ref{sec:graph-classifiers} and \ref{sec:graph-xai}, we define the notation used throughout the paper while describing the building blocks of this manuscript: Graph Neural Networks, the WL coloring algorithm, graph explainers and their evaluation. In Section~\ref{sec:method}, we detail our main contribution and introduce the \textsc{OpenGraphXAI} benchmark suite. In Section~\ref{sec:experiments}, we present a use case of the benchmark for experimenting with different graph explainers and evaluate them with robust statistics. Finally, in Section~\ref{sec:conclusion}, we draw our conclusions and highlight future research directions.

\section{Neural networks for graph classification}
\label{sec:graph-classifiers}

We define \emph{graphs} as tuples $\graph = (\nodeset_\graph,\edgeset_\graph)$, where $\nodeset_\graph=\{v_i : i \in 1, ..., |\nodeset_\graph| \}$ is the set of nodes and $\edgeset_\graph=\{(u,v) : u,v \in \nodeset_\graph\}$ is the set of (possibly directed) edges linking the nodes. A typical example of graph-structured data is molecules, where nodes are atoms and edges are chemical bonds between them.
We restrict our attention to \emph{attributed} (or \emph{labeled}) \emph{graphs}, where each node $v$ has an input label $x_v \in \mathcal{X}$ with $\mathcal{X}$ being a finite set, e.g., the set of atomic elements in the case of molecular graphs.
A \emph{subgraph} of $\graph$ is a graph $\graph' = (\nodeset_{\graph'},\edgeset_{\graph'})$ such that $\nodeset_{\graph'} \subseteq \nodeset_\graph$, $\edgeset_{\graph'} \subseteq \{ (u,v) \in \edgeset_{\graph} : u, v \in \nodeset_{\graph'}\}$ and $x_v' = x_v,\, \forall v \in \nodeset_{\graph'}$.
We use the notation $\graph' \sqsubseteq \graph$ to indicate that $\graph'$ is a subgraph of $\graph$.
Among all possible subgraphs of $\graph$, \emph{motifs} exhibit structural patterns of particular interest (e.g., they may correspond to explanatory patterns, as described later).
Finally, we define $d_\graph(v,u)$ as the shortest-path distance from $v$ to $u$, and $\Ego_\graph^{(\ell)}(v) = \{ u \in \mathcal{V} : d_\graph(v,u) \leq \ell \}$ as the set of nodes that belong to the $\ell$-radius ego-graph of $v$.

\subsection{Graph Neural Networks}
\label{sec:gnns}

Graph classification consists of learning an unknown function $f: \mathcal{G} \to \mathcal{Y}$ mapping graphs from an input domain $\mathcal{G}$ to the set of discrete labels $\mathcal{Y}$ denoting two or more classes.
For example, $\mathcal{Y} = \{0, 1\}$ in the case of binary classification tasks.
Borrowing from the chemical example above, class labels might indicate e.g. whether the molecule corresponding to the input graph is toxic or not.

Graph Neural Networks (GNNs) are a class of neural networks that take as input graph-structured data and return as output probabilistic predictions over the target domain.
In the context of graph classification, GNNs implement a parametric function $f_\Theta: \mathcal{G} \to \mathbb{R}^{|\mathcal{Y}|}$.
The output scores $\hat{\mathbf{y}}_\graph= f_\Theta(\graph)$ can be interpreted as probabilities (e.g. via softmax), with $\hat{y}_\graph = \argmax f_\Theta(\graph)$ as the predicted class.
Learning to approximate $f$ with $f_\Theta$ requires a labeled graph dataset $\mathcal{D} = \{(\graph, y_\graph) : \graph \in \mathcal{G}, y_\graph \in \mathcal{Y}\}$, and consists of optimizing the parameters $\Theta$ via gradient descent to minimize the cross-entropy between the GNN predictions for the input graphs and their corresponding labels.
GNNs for graph classification typically follow a modular architecture composed of three main components: (i) one or more \textit{message-passing} (MP) layers, (ii) a global node aggregation module, and (iii) a readout classifier.

The MP layers iteratively update node embeddings by exchanging and aggregating information among neighboring nodes, jointly encoding the local structure and the input node label.
Formally, the MP mechanism can be expressed by the following general iterative scheme (below, $\ell > 0$ indicates the MP layer or iteration):
\begin{subequations}\label{eq:mp-blueprint}
\label{eq:mp}
\begin{align}
    \mathbf{h}_v^{(0)} &= \mathbf{x}_v,\\
    \mathbf{h}^{(\ell)}_v &= \Upd\left(\mathbf{h}_v^{(\ell-1)},\, \Agg \left(\,\Multiset{\Msg(\mathbf{h}^{(\ell-1)}_v,\; \mathbf{h}^{(\ell-1)}_u) :u \in \NbrV}\,\right)\right),
\end{align}
\end{subequations}
where $\mathbf{h}_v^{(\ell)}$ denotes the embedding of node $v$ at layer $\ell$, and $\mathbf{x}_v$ is a vector representation of the input node label $x_v$ (e.g., its one-hot encoding).
The neighborhood of node $v$ is defined as $\mathcal{N}_v = \{u \in \mathcal{V} : (u,v) \in \edgeset_\graph\}$.
The functions $\Msg$, $\Agg$, and $\Upd$ respectively define the message computation, aggregation, and update operations: $\Msg$ computes pairwise messages between node $v$ and each of its neighbors; $\Agg$ aggregates these messages using a permutation-invariant operator such as sum, mean, or max pooling; and $\Upd$ updates the node representation by combining it with the aggregated information.
Different parameterizations of $\Msg$, $\Agg$, and $\Upd$ give rise to the various MP formulations proposed in the literature \citep{Wu2021}.

\begin{figure}
    \begin{center}
        \includegraphics[width=.75\textwidth]{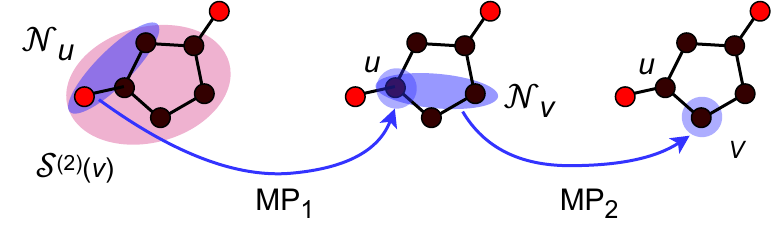}
    \end{center}
    \caption{The schematics of message-passing (MP) in GNNs. The representation $\mathbf{h}_v^{(2)}$ of node $v$ at layer $\ell = 2$ is obtained by aggregating and combining the representations $\left\{\mathbf{h}_u^{(1)} : u \in \NbrV\right\}$, which in turn are obtained by the same mechanism. After the two MP layers $\text{MP}_1$ and $\text{MP}_2$, all nodes within $2$ hops have contributed to the context of node $v$, that is $\Ego_\graph^{(2)}(v)$.}
    \label{fig:mp}
\end{figure}

Multiple MP layers enable node embeddings to progressively integrate information from increasingly distant nodes, thereby enhancing the model’s ability to learn contextual representations tailored for downstream graph-level predictions.
Precisely, the representation $\mathbf{h}^{(\ell)}_v$ of node $v$ has directly or indirectly received contributions from all nodes within distance $\ell$ of $v$, meaning its receptive field is $\Ego_\graph^{(\ell)}(v)$ \citep{Micheli}.

After $L > 0$ MP layers, the global aggregation or \emph{pooling} module combines the embeddings of all nodes within a graph into a single graph-level representation with a permutation-invariant operator, namely:
\begin{equation}
\label{eq:pool}
    \mathbf{h}_\graph = \Pool\left( \Multiset{\mathbf{h}^{(L)}_v : v \in \nodeset} \right).
\end{equation}
Finally, the global embedding is passed to a readout classifier (typically a linear layer or a multi-layer perceptron), which outputs class probabilities over the target domain. This architectural modularity allows GNNs to flexibly adapt to any graph classification task while exploiting the graphs' structural inductive biases.

\subsection{Expressiveness of GNNs}
\label{sec:express}

A pair of labeled graphs $\graph, \graph'$ is defined to be \emph{isomorphic}, denoted by $\graph \simeq \graph'$, if and only if $\nodeset_\graph = \nodeset_{\graph'}$, $\edgeset_\graph = \edgeset_{\graph'}$, and $x_v = x'_v$ for all nodes $v$.
As isomorphic graphs are indistinguishable, any function on the domain of graphs $f: \mathcal{G} \to \mathcal{Y}$ must satisfy $\graph \simeq \graph' \implies f(\graph) = f(\graph')$.
Which graphs can be distinguished by a message-passing GNN characterizes the model's \emph{expressiveness}, as a GNN must be able to learn different graph representations $\mathbf{h}_\graph \neq \mathbf{h}_{\graph'}$ for different graphs $\graph \not\simeq \graph'$ with different targets $f(\graph) \neq f(\graph')$.
\cite{Morris2019,Xu2019} have established that the class of GNNs described in the previous section can be at most as expressive as the WL isomorphism test, which is able to distinguish just a subset of all non-isomorphic graphs.

\subsubsection{WL node coloring}
\label{sec:wl}

The Weisfeiler--Leman (WL) node coloring algorithm \citep{Weisfeiler1968,Morris2023} has been widely employed as a heuristic test of graph isomorphism thanks to its computational complexity of $O\left(|\mathcal{E}| \log |\mathcal{V}|\right)$.
\begin{figure}
    \begin{center}
        \includegraphics[width=.8\textwidth]{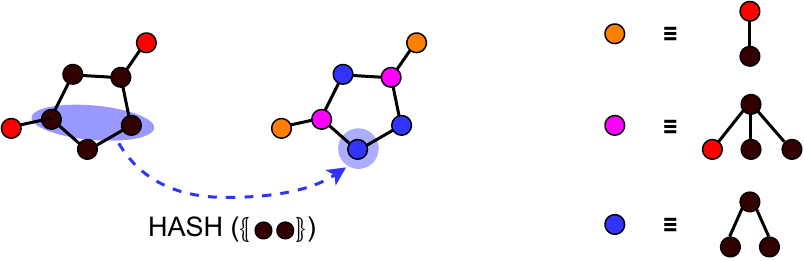}
    \end{center}
    \caption{The schematics of WL coloring. The blue color assigned to the node highlighted in the rightmost graph is obtained by hashing the color of its immediate neighbors at the previous iteration (leftmost graph). Each WL color identifies a unique unfolding tree (right).}
    \label{fig:wl}
\end{figure}
Starting from an initial set of colors $\mathcal{C}^{(0)} = \mathcal{X}$ and a node-wise color assignment $\textsf{c}^{(0)}_v = x_v,\, \forall v \in \nodeset_\graph$, each WL iteration $\ell > 0$ assigns a new unique color $\textsf{c}^{(\ell)}_v \in \mathcal{C}^{(\ell)}$ to each node based on the multi-set of neighbor labels, as illustrated in Fig.~\ref{fig:wl}:
\begin{subequations}
\label{eq:wl}
\begin{align}
    \textsf{c}_v^{(0)} &= x_v,\\
    \textsf{c}^{(\ell)}_v &= \Hash\left( \Multiset{\textsf{c}^{(\ell-1)}_u : u \in \mathcal{N}_v} \right)
\end{align}
\end{subequations}
In other words, two nodes are assigned the same WL color if and only if they share the same multi-set of neighbor colors at the previous iteration.
More formally, a WL color $\mathsf{c}_v^{(\ell)}$ identifies the WL-equivalence class of rooted subgraphs of radius $\ell$ that are indistinguishable from the one rooted at $v$.
Therefore, if two nodes $v, u$ are assigned equal colors, i.e., $\textsf{c}^{(\ell)}_v = \textsf{c}^{(\ell)}_u$, their unfolding trees of depth $\ell$ are equal \citep{Kriege2022,DInverno2024}.
We define $\mathcal{C}^{(\ell)}_\graph = \{\mathsf{c}_v^{(\ell)} : v \in \nodeset_\graph \}$ as the set of all WL colors at iteration $\ell$.

The color assignment induces a partition over the graph nodes $\wp^{(\ell)}_\graph = \{\Psi_\graph(\mathsf{c}) : \mathsf{c} \in \mathcal{C}^{(\ell)}_\graph\}$, where each node belongs to a unique $\Psi_\graph(\mathsf{c}) = \{v \in \nodeset : \mathsf{c}_v^{(\ell)} = \mathsf{c}\}$.
A finite number of iterations $L < |\nodeset|$ of Eq.~\eqref{eq:wl} is sufficient for the coloring to be stable, that is for $\wp^{(L+1)}_\graph = \wp^{(L)}_\graph$ \citep{Kiefer2020}.
After such $L > 0$ steps, the graph is represented as the histogram of its node colors
\begin{equation}
\label{eq:wl-pool}
    \textsf{h}_\graph = \Hash\left( \Multiset{\textsf{c}^{(L)}_v : v \in \nodeset_\graph} \right)
\end{equation}

Two graphs $\graph, \graph'$ such that $\textsf{h}_\graph = \textsf{h}_{\graph'}$ are considered equivalent by the WL test, denoted by $\graph \simwl \graph'$.
As the WL test is a heuristic for graph isomorphism, it holds that $\graph \not\simwl \graph' \implies \graph \not\simeq \graph'$, but the converse is not true.

\subsubsection{Learnable decision functions}
\label{sec:learnable}

Both message-passing GNNs and the WL coloring algorithm rely on node representations refined by neighborhood aggregation, as can be noticed by the similarity of Eq.~\eqref{eq:mp} with Eq.~\eqref{eq:wl}.
Indeed, \cite{Xu2019} have proved that the WL isomorphism test constitutes the upper bound of expressivity for this class of GNNs, that is $\mathbf{h}_\graph \neq \mathbf{h}_{\graph'} \implies \graph \not\simwl \graph'$.
This limit of expressiveness ($\mathbf{h}_\graph \neq \mathbf{h}_{\graph'} \Leftrightarrow \graph \not\simwl \graph'$) is attained when GNNs learn injective functions in Eq.~\eqref{eq:mp}--\eqref{eq:pool}.

Due to their WL expressiveness, GNNs can only learn functions $f_\Theta$ that preserve the equivalence of unfolding trees \citep{Scarselli2009comp,DInverno2024}.
In other words, GNNs for graph classification can only make their decision based on occurrences of WL colors in the graph \citep{Barcelo2020}.
In particular, a GNN with $L$ message-passing layers can distinguish graphs up to the occurrence colors in $\mathcal{C}^{(L)}_\graph$ \citep{DInverno2024}.
This result provides a characterization of the type of sub-structure patterns that can serve as explanations for class attributions by message-passing GNNs.
Namely, rooted subgraphs consisting of nodes $\Ego_\graph^{(L)}(v)$ corresponding to the root node color $\textsf{c}^{(L)}_v \in \mathcal{C}^{(L)}_\graph$.

\section{Graph-XAI}\label{sec:graph-xai}

\subsection{Graph explainers}
\label{sec:graph-explainers}

This work focuses on \textit{instance-based} graph explainers \citep{TaxSurvey}, which aim at identifying graph components (nodes, edges, or features) that contribute most to the model's prediction on a specific input graph $\graph$. Among these, \textit{node-centric} graph explainers work by assuming that there exists a Ground Truth (GT) motif $\motif_\graph = (\nodeset_{\motif_\graph}, \edgeset_{\motif_\graph}) \sqsubseteq \graph$ within the graph, whose presence determines the prediction. To represent the GT motif compactly, a \textit{GT mask} vector $\gtruth_\graph \in \{0,1\}^{|\nodeset_\graph|}$ can be computed, with entries defined as follows:
\begin{equation}
\gtruth_\graph[i] = \mathbbm{1}\left[ v_i \in \nodeset_{\motif_\graph} \right].
\end{equation}
Accordingly, this class of explainers seeks to recover the GT motif by outputting an \textit{explanation mask} $\mask_\graph$, i.e., a vector where the $i$-th entry quantifies the predicted degree of membership of node $v_i$ in the GT motif. We distinguish between \textit{soft} explanation masks $\mask_\graph \in \mathbb{R}^{|\nodeset_\graph|}$, with real-valued scores, and \textit{hard} explanation masks $\mask_\graph \in \{0,1\}^{|\nodeset_\graph|}$, which specifically indicate whether each entry is a node of the predicted explanatory subgraph. Notice that any soft explanation mask can be converted into a hard one by appropriate thresholding (e.g., selecting the top-$K$ scoring nodes).

Graph explainers are evaluated using specific benchmark datasets where labeled examples are paired with the set of nodes composing the GT motif. More formally, a graph-XAI dataset is a collection $\xaidataset = \{\left(\graph, y_\graph, \nodeset_{\motif_\graph} \right) : \graph \in \mathcal{G}, y_\graph \in \mathcal{Y}\}$. Therefore, an explainer can be evaluated by checking the degree of matching between the predicted mask and the GT motif mask, averaged across the entire dataset:

\begin{equation}
    \frac{1}{|\xaidataset|} \sum_{(\graph, y_\graph, \nodeset_{\motif_\graph}) \in \xaidataset} \delta(\gtruth_\graph, \mask_\graph),
\end{equation}
where $\delta$ is any metric that quantifies the alignment between the masks (e.g., intersection over union for hard 
explanation masks).

\subsection{Related works on graph-XAI evaluation}

The assessment of graph explainers \citep{TaxSurvey,survey2XAI} relies on evaluating the quality of the explanations they yield. Since the inception of the first explainers for GNNs \citep{Baldassarre,Pope,gnnexplainer,pgExplainer}, the development of appropriate metrics and benchmarks has been a key focus in the field. Metrics to evaluate graph explainers include unsupervised ones such as fidelity \citep{Pope,amara2022graphframex,longa2025explaining}, which quantifies the degree to which the explanation reflects the GNN's output, or supervised ones like plausibility \citep{rathee2022bagel}, which measures how well the explanation aligns with the GT explanation associated with a given class. Other metrics cover additional desiderata such as sparseness (succinctness) and connectedness (compactness) of the explanations.

Benchmark datasets with integrated GTs for the evaluation of graph explainers have predominantly been of synthetic nature, typically by augmenting randomly generated Barabási-Albert (BA) graphs with explanatory motifs. For example, in the BA2Grid dataset \citep{pgExplainer}, positive samples are BA graphs with a grid graph motif attached, while negative samples are random BA graphs without an attached motif. Among other synthetic datasets, we mention BA2Motif, GridHouse, HouseColors, and Stars \citep{longa2025explaining}. However, these datasets suffer from key limitations \citep{faber2021comparing}. First of all, the explanatory patterns in these datasets are synthetic motifs that are quite regular and typically different from real-world, copmplex patterns. Furthermore, \cite{fontanesi2024xai} discovered that despite being trained to perfect accuracy, different GNN architectures rely on different, but equally valid, explanations than the original GTs.

Real-world graph-XAI datasets have been introduced by \citet{NEURIPS2020_google} and later included and extended in the GraphXAI library \citep{agarwal2023evaluating}. These datasets are based on chemical molecules from known data sources \citep{Mutagenicity1, Mutagenicity2, ZINC15}, and are designed for binary graph classification tasks. They include a variant of Mutagenicity \citep{Mutagenicity2}, and three subsets of the ZINC15 database \citep{ZINC15}, namely FluorideCarbonyl, AlkaneCarbonyl, and Benzene. Their GT explanations have been established by leveraging chemical domain expertise. Specifically, the Mutagenicity dataset distinguishes molecules based on the presence of known toxicophores such as \texttt{NH$_{2}$}, \texttt{NO$_{2}$}, aliphatic halide, nitroso, and azo-type. The FluorideCarbonyl and AlkaneCarbonyl datasets classify molecules based on the co-occurrence of a fluoride and carbonyl or an alkane and carbonyl functional group, respectively. Lastly, the Benzene dataset distinguishes molecules containing at least one benzene ring from those that lack one. In contrast, the dataset proposed by \cite{varbella2024powergraph} features GT explanations for cascading failure analysis in power grids, which have been derived through computer model simulations. In general, datasets with GT explanations based on real-world graphs are difficult to generate, as identifying the GT motifs associated with the classes requires the knowledge and effort of a domain expert. This complexity explains their limited availability in the literature.

All the datasets mentioned above are routinely collected into benchmarks by common graph-XAI libraries \citep{agarwal2023evaluating, amara2022graphframex, rathee2022bagel, varbella2024powergraph}.
Our approach to overcoming the limitations of current practices for dataset collation in graph-XAI is orthogonal and leverages an algorithm to extract synthetic explanatory subgraphs from curated subsets of real-world datasets. Therefore, our approach has the distinctive advantage that GT explanations are derived from the same structural patterns as the originating graph datasets.

\section{Methods}
\label{sec:method}

From the discussion in Section~\ref{sec:graph-explainers}, it follows that any graph classification dataset can be transformed into a graph-XAI evaluation dataset by augmenting the labeled samples with the corresponding GT masks. 
In this section, we describe a method to automate this process, based on two intuitive steps:
\begin{enumerate}
    \item First, mine a large set of sub-graph motifs of different sizes, balancing exhaustiveness and computational cost.
    \item Then, generate a collection of suitable XAI benchmarks from the set of potentially discriminating motifs.
\end{enumerate}
Our method can be applied to any graph classification dataset with discrete input features, making it particularly suitable e.g. in the case of chemical data.

\subsection{Automating the construction of XAI benchmarks}
\label{sec:method-details}

Recall that a graph classification dataset is a set of labeled graphs $\mathcal{D} = \{(\graph, y_\graph) : \graph \in \mathcal{G}, y_\graph \in \mathcal{Y} \}$. We restrict our focus to datasets for binary graph classification where the target domain contains only two class labels, i.e., $\mathcal{Y} = \{0, 1\}$. Our objective is to construct an XAI benchmark dataset $\xaidataset = \{(\graph, y_\graph, \nodeset_{\motif_{\graph}}) : (\graph, y_\graph) \in \mathcal{D}'\}$ with $\mathcal{D}' \subseteq \mathcal{D}$, where the GT motifs $\motif_\graph \sqsubseteq \graph$ are guaranteed to occur (or never to occur) in graphs with a certain class label. 

To avoid the computational burden of testing for subgraph isomorphism, we resort to working with WL colors (and thus, with the corresponding rooted subgraphs) rather than arbitrary subgraphs. More precisely, the discriminating motifs are extracted from the  set:
\begin{equation}\label{eq:wl-dataset-colors}
\mathcal{C} = \bigcup_{\graph \in \mathcal{D}} \mathcal{C}_{\graph}^{(\leq L)}, \qquad \text{where} \quad \mathcal{C}_{\graph}^{(\leq L)} = \bigcup_{1 \leq \ell \leq L} \mathcal{C}_{\graph}^{(\ell)},
\end{equation}
holding all possible colors extracted by applying $L$ iterations of WL coloring over the graphs in $\mathcal{D}$. Note that $\mathcal{C}^{(\ell)} \subseteq \mathcal{C}$ identifies the subset of all colors found at the $\ell$-th WL iteration in $\mathcal{C}$. This restriction allows us to effectively side-step the subgraph isomorphism problem, since given a color $\mathsf{c} \in \mathcal{C}$, the occurrence of $\mathsf{c} \in \mathcal{C}_{\graph}^{(\leq L)}$ tells us that $\graph$ contains a WL-equivalent rooted subgraph.
Furthermore, this choice ensures that the discriminating motifs $\motif_\graph$ are within the range of expressiveness of GNNs, as discussed in Section~\ref{sec:express}.

\begin{figure}[t!]
    \centering
    \includegraphics[width=\textwidth]{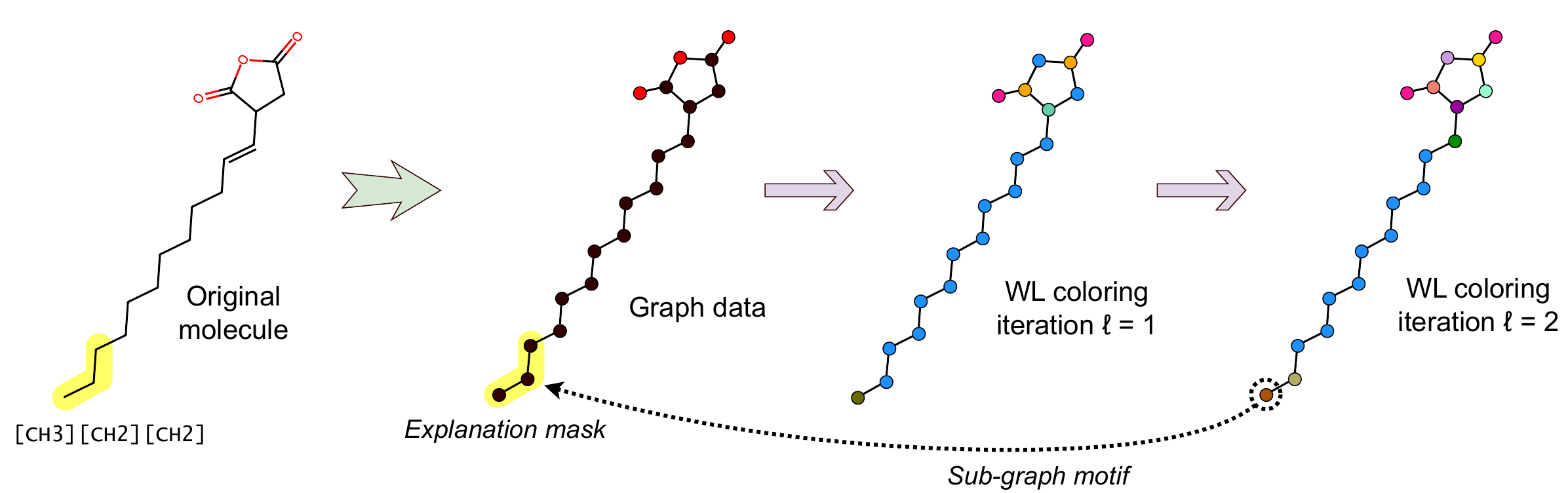}
    \caption{We show the mechanics of the proposed algorithm for automatically generating a GT-annotated benchmark from any graph classification dataset through an exemplary molecule represented as a graph. WL coloring is used to mine the set of sub-graph motifs available for the construction of our XAI benchmarks. A motif corresponding to the WL label $\textsf{m} \in \mathcal{C}_2$ has been chosen. The explanation mask is constituted by the $2$-radius ego-graphs of nodes with WL color $\textsf{m}$. In this example, the mined motif can be traced back to the fragment \texttt{[CH3][CH2][CH2]}.}
    \label{fig:wl-coloring}
\end{figure}

At this point, we mine $\mathcal{C}$ to find WL colors that discriminate graph pairs $(\graph, y_\graph) \in \mathcal{D}$ according to two different policies:
\begin{description}
    \item[Case 1:] we search for one discriminating color $\bar{\mathsf{c}}$ such that
    \begin{equation}
        \begin{cases}
        \bar{\mathsf{c}} \notin \mathcal{C}^{(\leq L)}_\graph & \iff  \;y_\graph = 1-y\\
        \bar{\mathsf{c}} \in \mathcal{C}^{(\leq L)}_\graph & \iff \;y_\graph = y,
        \end{cases}                                 
    \end{equation}
    i.e., it only appears in graphs with class label $y$.
    \item[Case 2:] we search for two discriminating colors $\bar{\mathsf{c}}_0$ and $\bar{\mathsf{c}}_1$ such that: 
    \begin{equation}
        \begin{cases}
            \bar{\mathsf{c}}_0 \in \mathcal{C}^{(\leq L)}_\graph \,\land\, \bar{\mathsf{c}}_1 \notin \mathcal{C}^{(\leq L)}_\graph & \iff  \;y_\graph = 0\\
            \bar{\mathsf{c}}_1 \notin \mathcal{C}^{(\leq L)}_\graph \,\land\, \bar{\mathsf{c}}_1 \in \mathcal{C}^{(\leq L)}_\graph & \iff  \;y_\graph = 1,
        \end{cases}                                    
    \end{equation}
    i.e., each color occurs only in graphs with a certain class label.
\end{description}
These policies allow us to construct the set $\mathcal{D}'$ in different ways, by filtering out all graphs that do not respect the chosen case rules. Then, we proceed to constructing the explanation motif $\motif_\graph$ for all $\graph \in \mathcal{D}'$.
In particular, assuming to be working with the discriminating WL color $\bar{\textsf{c}} \in \mathcal{C}^{(\ell)}$ of Case 1 (Case 2 is defined analogously), the explanation motif $\motif_\graph = (\nodeset_{\motif_{\graph}}, \edgeset_{\motif_{\graph}}) \sqsubseteq \graph$ is a subgraph whose nodes are those of the union graph comprising all rooted subgraphs in $\graph$ corresponding to $\bar{\mathsf{c}}$-colored nodes $\Psi_\graph(\mathsf{c}) \subset \nodeset$. In practice:
\begin{align}
\nodeset_{\motif_{\graph}} &= \bigcup_{v \in \Psi_\graph(\bar{\mathsf{c}})} \Ego_\graph^{(\ell)}(v),\\
\edgeset_{\motif_{\graph}} &= \{(u, v) \in \edgeset_\graph : u, v \in \nodeset_{\motif_\graph} \}.
\end{align}
This process is illustrated in Figure~\ref{fig:wl-coloring}.

Since considering all possible WL colors (Case 1) and color pairs (Case 2) is still computationally onerous, we propose a heuristic approach to choose discriminating WL colors based on their frequencies in the respective target classes.
Let:

\begin{equation}\label{eq:wl-class-freq}
    \freq_y(\textsf{c}) = \sum_{(\graph, y_\graph) \in \mathcal{D}} \mathbbm{1}[\textsf{c} \in \mathcal{C}_\graph^{(\leq L)} \land y_\graph = y]
\end{equation} 
be the number of graphs of class $y$ where $\textsf{c} \in \mathcal{C}$ occurs, and let $\Delta\!\freq(\textsf{c}) = \freq_1(\textsf{c}) - \freq_0(\textsf{c})$.
The best candidate as discriminating WL colors for class $1$ (resp. class $0$) would be the $\textsf{c}$'s with largest (resp. smallest) $\Delta\!\freq(\textsf{c})$, as they are the most skewed towards one of the classes.
Once the candidate colors have been selected, the XAI benchmark $\mathcal{D}_\text{XAI} = \left\{\left(\graph, y_\graph, \nodeset_{\motif_\graph}\right)\right\}$ is generated by collecting graphs that satisfy the color-class assignment, along with the GT masks obtained via the ego sub-trees.
The overall procedure is detailed in Algorithm~\ref{alg:xai-generator}.

\begin{algorithm}[h!]
\caption{Generate graph-XAI datasets from real-world graph datasets}
\label{alg:xai-generator}
\begin{algorithmic}[1]

    \Require Binary graph classification dataset $\mathcal{D}$, number of WL iterations $L > 0$, maximum number of candidate WL labels $K > 0$

    \linespread{1.2}\selectfont
    
    \State Compute per-graph WL colors $\mathcal{C}^{(\leq L)}_\graph = \bigcup_{1 \leq \ell \leq L} \mathcal{C}_{\graph}^{(\ell)}$ \Comment{Equation \ref{eq:wl}}
    
    \State Compute per-dataset WL colors $\mathcal{C}$ \Comment{Equation \ref{eq:wl-dataset-colors}} 

    \State Compute per-class color frequencies $\freq_y(\textsf{c}),\;\forall \textsf{c} \in \mathcal{C}$  \Comment{Equation \ref{eq:wl-class-freq}}
    
    \State $\hat{\mathcal{C}}_0 \gets \left\{ \textsf{c} \in \mathcal{C} : \textsf{c} \in \textsc{top}_K(-\Delta\!\freq(\textsf{c})) \right\}$
    
    \State $\hat{\mathcal{C}}_1 \gets \left\{ \textsf{c} \in \mathcal{C} : \textsf{c} \in \textsc{top}_K(\Delta\!\freq(\textsf{c})) \right\}$
    
    \For{$y \in \{0, 1\}$} \Comment{Case 1 policy}
    \For{$\bar{\mathsf{c}} \in \hat{\mathcal{C}}_y$}
    \State $\mathcal{D}_y \gets \left\{\left(\graph, y, \bigcup_{v \in \Psi_\graph(\bar{\mathsf{c}})} \Ego_\graph^{(\ell)}(v)\right) : (\graph, y) \in \mathcal{D},\, \bar{\mathsf{c}} \in \mathcal{C}^{(\leq L)}_\graph \right\}$
    \State $\mathcal{D}_{1-y} \gets \left\{\left(\graph, 1-y, \varnothing\right) : (\graph, 1-y) \in \mathcal{D},\, \bar{\mathsf{c}} \notin \mathcal{C}^{(\leq L)}_\graph \right\}$
    \State \textbf{yield} $\mathcal{D}_\text{XAI} = \mathcal{D}_y \cup \mathcal{D}_{(1-y)}$
    \EndFor
    \EndFor
    
    \For{$\bar{\mathsf{c}}_0, \bar{\mathsf{c}}_1 \in \hat{\mathcal{C}}_0 \times \hat{\mathcal{C}}_1$} \Comment{Case 2 policy}
    
    \State $\mathcal{D}_0 \gets \left\{\left(\graph, 0, \bigcup_{v \in \Psi_\graph(\bar{\mathsf{c}}_0)} \Ego_\graph^{(\ell)}(v)\right) : (\graph, 0) \in \mathcal{D},\, \bar{\mathsf{c}}_0 \in \mathcal{C}^{(\leq L)}_\graph,\, \bar{\mathsf{c}}_1 \not\in \mathcal{C}^{(\leq L)}_\graph\right\}$

    \State $\mathcal{D}_1 \gets \left\{\left(\graph, 1, \bigcup_{v \in \Psi_\graph(\bar{\mathsf{c}}_1)} \Ego_\graph^{(\ell)}(v)\right) : (\graph, 1) \in \mathcal{D},\, \bar{\mathsf{c}}_0 \not\in \mathcal{C}^{(\leq L)}_\graph,\, \bar{\mathsf{c}}_1 \in \mathcal{C}^{(\leq L)}_\graph\right\}$
    \State \textbf{yield} $\mathcal{D}_\text{XAI} = \mathcal{D}_0 \cup \mathcal{D}_1$
    \EndFor
    
\end{algorithmic}
\end{algorithm}

\subsection{The \textsc{OpenGraphXAI} benchmark suite}

We apply our method to create \textsc{OpenGraphXAI}, a curated collection of 15 novel XAI benchmarks for graph classification, listed in Table~\ref{tab:tasks}.
All the datasets of our benchmarking suite are based on real-world molecular graphs derived from the TUDataset collection by \cite{Morris2020}.
They have been constructed to cover multiple GT-availability scenarios while offering mostly class-balanced tasks to solve.
As part of the collection, we provide tasks featuring GT explanations for both classes of equal-size sub-graph motifs (\texttt{alfa}, \texttt{delta}) or of different sizes (\texttt{bravo}, \texttt{charlie}, \texttt{echo}), as well as datasets providing GT explanations for only one of the two classes (\texttt{foxtrot} to \texttt{oscar}).
Visual examples of the different GT masks across the datasets are illustrated in Appendix~\ref{appx:ogx-gts}.
We also provide suggested scaffold splits into $70\%$ training, $20\%$ validation, and $10\%$ test sets for each benchmark, with samples stratified by class and graph size.
As supplementary material, we make available the WL labels to generate an extended collection of 2000+ benchmarks from other TU datasets.
We stress that GT labels, even in the case of the original datasets being molecules, do not necessarily correspond to molecular fragments or other chemically-interpretable substructures, but only to subgraph patterns distinguishable by WL-expressive GNNs.

\begin{table}[t!]
	\caption{Datasets in the \textsc{OpenGraphXAI} benchmark suite. The Balance column reports the ratio between the minority class and the majority class.}
    \vspace{\belowcaptionskip} 
	\label{tab:tasks}
	\centering
	\begin{tabular}{llcccccc}
		\toprule
		& & \multicolumn{2}{c}{\textbf{WL iterations}} & \multicolumn{3}{c}{\textbf{Samples}} & \\
		\cmidrule(r){3-4} \cmidrule(r){5-7}
		\textbf{Task} & \textbf{Original dataset} & Class 0 & Class 1 & Class 0 & Class 1 & Total & \textbf{Balance} \\
		\midrule
		\texttt{alfa}     & NCI1          & 3 & 3 & 505  & 598  & 1103 & 0.84 \\
		\texttt{bravo}    & NCI1          & 3 & 1 & 573  & 551  & 1124 & 0.96 \\
		\texttt{charlie}  & NCI1          & 0 & 3 & 463  & 543  & 1006 & 0.85 \\
		\texttt{delta}    & NCI109        & 3 & 3 & 515  & 563  & 1078 & 0.91 \\
		\texttt{echo}     & NCI109        & 3 & 2 & 598  & 558  & 1156 & 0.93 \\
		\texttt{foxtrot}  & NCI109        & --- & 2 & 1195 & 987  & 2182 & 0.83 \\
		\texttt{golf}     & TOX21\_AHR    & 4 & --- & 833  & 962  & 1795 & 0.87 \\
		\texttt{hotel}    & TOX21\_ER-LBD & 4 & --- & 444  & 457  & 901  & 0.97 \\
		\texttt{india}    & TOX21\_P53    & 4 & --- & 658  & 590  & 1248 & 0.90 \\
		\texttt{juliett}  & TOX21\_ER-LBD & 5 & --- & 563  & 455  & 1018 & 0.81 \\
		\texttt{kilo}     & MCF-7         & 3 & ---  & 2007 & 1978 & 3985 & 0.99 \\
		\texttt{lima}     & MOLT-4        & 3 & --- & 3086 & 2786 & 5872 & 0.90 \\
		\texttt{mike}     & P388          & 3 & --- & 2444 & 2225 & 4669 & 0.91 \\
		\texttt{november} & PC-3          & 2 & --- & 1609 & 1531 & 3140 & 0.95 \\
		\texttt{oscar}    & SW-620        & 3 & --- & 2340 & 2273 & 4613 & 0.97 \\
		\bottomrule
	\end{tabular}
\end{table}

\subsection{Applicability, limitations, and extension of our framework}

As stated in Section~\ref{sec:method-details}, our method for the construction of XAI benchmarks is restricted to GT motifs that can be distinguished via WL coloring, which in turn can be applied only to graphs with discrete input features.
This restriction is necessary to avoid the combinatorial explosion of enumerating all possible sub-graphs, and the computational complexity of the general sub-graph isomorphism problem.
While our method can be straightforwardly extended to generate multi-class tasks, we only focus on binary classification, which is the typical setting of previous real-world XAI benchmarks.
We also generate at most a single GT motif per class, a limitation we plan to address in the future by considering general predicates on WL colors as GT explanations, e.g. from the ALCQ description logic \citep{Barcelo2020}.

An advantage of our method is that the classification rules of the generated datasets $\xaidataset$ are ensured to be learnable by message-passing GNNs by construction, as the corresponding discriminating motifs are within the range of expressiveness of this class of GNNs (see Sec.~\ref{sec:express}).
However, types of neural models for graphs that depart from the message-passing approach may not be modeled by the expressiveness of the WL test as described in Section~\ref{sec:wl} \citep{Zhang2025}.
For example, higher-order GNNs are as expressive as the $k$-WL isomorphism test \citep{Morris2019}, Graph Transformers with structural positional encodings as the SEG-WL test \citep{Zhu2023}, Path-GNNs as the PATH-WL test \citep{Graziani2024}, and so on.
In those cases, our method can be readily extended to cover the differing classes of learnable motifs by replacing the ``classic'' WL coloring in Algorithm~\ref{alg:xai-generator} with the corresponding extension of WL that matches the model's expressiveness.

\section{Evaluating graph explainers: A use case}
\label{sec:experiments}

To showcase the contribution that a large collection of benchmarks such as ours can make to the research in graph-XAI, we perform an experimental evaluation of a set of explainer methods for graph classification.
Specifically, we focus on \textit{post-hoc}, node-centric, and local explainers applied to a trained GNN model that achieves a high performance across \textsc{OpenGraphXAI}'s classification tasks, and measure how well the explanations align with the GT masks.

\subsection{Experiments}

\subsubsection{Model to explain}

Below, we describe the architecture of the model whose predictions are analyzed by the different graph explainers.

As anticipated in Section~\ref{sec:gnns}, the architecture of the graph classifier follows a modular setup composed of a stack of MP layers that produce node embeddings, followed by an aggregation step that produces a graph embedding, which are used by a downstream readout layer to produce the final prediction. Here, we document the design choices of this general architecture for the specific use case.

For the MP layers, we choose the Graph Isomorphism Network (GIN) implementation. The rationale for choosing GIN is its wide use for graph classification tasks, and its proven expressivity. Indeed, GIN can theoretically achieve the upper bound of expressiveness of the WL isomorphism test \citep{Xu2019}, which ensures us that it is (in principle) capable of identifying the WL-structure corresponding to the explanation. In practice, GIN conforms to the MP paradigm (Eq.~\ref{eq:mp-blueprint}) by implementing \textsc{Msg} as a function returning the neighbor node embedding, \textsc{Agg} as sum, and \textsc{Upd} as a multi-layer perceptron (MLP) applied to the sum of the current node embedding and the embedding of the aggregated neighbor messages. More precisely, GIN updates a node embedding $\mathbf{h}_v^{(\ell-1)} \in \mathbb{R}^{d^{(\ell-1)}}$ to $\mathbf{h}_v^{(\ell)} \in \mathbb{R}^{d^{(\ell)}}$ (with $\ell > 0$) as follows:
\begin{equation}
    \mathbf{h}_v^{(\ell)} = \text{MLP}_{\theta^{(\ell-1)}} \left((1 + \epsilon)\; \mathbf{h}_v^{(\ell)} + \sum_{u \in \mathcal{N}_v} \mathbf{h}_u^{(\ell-1)}\right),
\end{equation}
where $\theta^{\ell} \in \mathbb{R}^{d^{(\ell)} \times d^{(\ell-1)}}$ are layer-wise learnable parameters with $d \in \mathbb{N}$ denoting embedding dimensions, and $\epsilon \in \mathbb{R}$ scales the importance of the current node embedding with respect to the neighbor embeddings. In our experiments, we set $\epsilon=0$.

For the node aggregation module, which computes the graph representation $\mathbf{h}_\graph \in \mathbb{R}^{d^{(L)}}$ after $L$ MP layers, we choose the sum operator, which is typical in graph classification tasks. Therefore, the graph embedding is computed as follows:
\begin{equation}\label{eq:pooling}
    \mathbf{h}_\graph = \sum_{v \in \nodeset_\graph} \mathbf{h}_v^{(L)}.
\end{equation}

Finally, the readout module consists of an affine transformation of the graph embedding, followed by a softmax function to transform the output into a probability distribution over the label space:
\begin{equation}\label{eq:logits}
    \hat{y}_\graph = \text{softmax}\left( \theta^{\text{(out)}} \mathbf{h}_\graph + b \right),
\end{equation}
where $\theta^{\text{out}} \in \mathbb{R}^{2 \times d^{(L)}}$ and $b \in \mathbb{R}^2$ are the learnable parameters of the affine transformation. 
The parameters $\Theta$ of the resulting GNN were trained end-to-end by minimizing the cross-entropy loss function between the class probability distribution predicted by the GNN and the graph label $y_\graph$.
During the experiments, we tuned the GNN architecture by selecting $L$ (the number of MP layers) and $d$ (the node embedding dimension). Furthermore, we tuned the learning rate and weight decay coefficients used for the gradient descent algorithm. Details on the hyper-parameter selection and the training performed for our experiments are reported in Appendix~\ref{appx:exp-details}.

\subsubsection{Graph Explainers}
We experiment with 4 local (node-centric) \textit{post-hoc} explainers. The rationale to select the explainers pool is their adoption by the graph-XAI community (essentially, we choose the most widely used). 

The \textbf{Saliency} method was initially applied to the image domain \citep{Simonyan14a} and then adapted for GNNs by \citet{Baldassarre} and \citet{agarwal2023evaluating}. It computes the entries of the importance mask for each graph node as follows:
\begin{equation}
    \mask_\graph[i] = \left| \frac{\partial\, f_{\Theta}(\graph)_{[y]}}{\partial\, \mathbf{x}_{v_i}} \right|,
\end{equation}
where $\text{GNN}(\graph)_{[y]}$ is is the GNN’s output score for class $y$ given graph $\graph$. In short, the importance is the sum of the gradients of the output class with respect of each node feature; the higher the score, the higher the importance of the associated node feature. 

\textbf{Integrated Gradients} (IntGrad) \citep{IntGrad} is a technique whose aim is to compute the contributions of the input features to the output prediction $y_\graph$ relatively to a baseline point $\mathbf{x}'$ which encodes the complete absence of information. Specifically, IG computes the path integral of the gradients along the straight line from the baseline $\mathbf{x'}$ to the input $\mathbf{x}$ as:
\begin{equation}
    \mask_\graph[i] = (\mathbf{x}_{v_i} - \mathbf{x}_{v_i}') \odot \int_{\alpha=0}^1 \frac{\partial\, f_\mathbf{\Theta}(\mathbf{x'} + \alpha \odot (\mathbf{x}-\mathbf{x'}))}{\partial\, \mathbf{x}_{v_i}}
\end{equation}
where $\alpha \in [0,1]$ is the interpolation coefficient of the path. Similarly for the Saliency method, the final node importance is computed by summing up the feature-wise importance scores.

\textbf{GNNExplainer} (GNNExpl) \cite{gnnexplainer} is the most widely used and referenced explainable AI technique for GNNs. It works by learning a soft mask over edges to identify the subgraph most influential for the GNN's prediction. The masks are optimized to maximize the mutual information between the prediction on the masked graph and the original prediction. Essentially, GNNExpl computes:
\begin{equation}
    \mask_\graph[i] = \sum_{j} M_{\edgeset_\graph}[i,j],
\end{equation}
where $M_{\edgeset_\graph}[i,j]$ is the entry in the learned edge mask $M_{\edgeset_\graph}$ containing the importance of edge $(v_i, v_j)$.
Node importances are then derived by summing the edge importances incident to each node. Here, we use a node-centric variant which learns soft mask over the nodes, as implemented in the PyTorch Geometric library \citep{Fey/Lenssen/2019}.

\textbf{Class Activation Mapping} (CAM) \citep{CAM,Pope} is a model-specific technique that requires the model to adopt a global add pooling layer before its readout component. Specifically, CAM rewrites the computation of the individual logits (Eqs. \ref{eq:pooling} and \ref{eq:logits}) as a sum of individual node contributions:
\begin{equation} \label{eq:cam}
	\mask_\graph[i] = \theta_{[y]} \sum_{v \in \nodeset_\graph} \mathbf{h}^{(L)}_{v} +b_{[y]} = \sum_{v \in \nodeset_\graph} \left(\theta_{[y]}\, \mathbf{h}^{(L)}_{v} + \frac{b_{[y]}}{|\nodeset_\graph|}\right)
\end{equation}
Above, each addend in the rightmost summation constitutes the importance score of a node as it quantifies a node's contribution to the logit value associated with class $y$.

Finally, to ensure the explanation computed by the graph explainers under scrutiny are not purely stochastic, we also evaluate a \textbf{Random} baseline that assigns node importances at random.

\subsubsection{Evaluation metric}
To measure the alignment between an explanation $\mask_\graph$ and a GT mask $\gtruth_\graph$ we adopt the \emph{plausibility} metric \citep{rathee2022bagel,longa2025explaining}, which is a standard metric designed to evaluate soft explanation masks. Specifically, plausibility is defined as the area under the ROC curve (AUROC) of the soft importance mask predicted by the explainer with respect to the GT mask.
We report the average plausibility score over all test graphs in each XAI benchmark for each of the five explainer methods, using the same evaluation framework described in Section~\ref{sec:graph-explainers}.

\subsubsection{Experimental setting}
For each dataset, we tuned the GIN-based architecture with hold-out model selection and a grid search algorithm to optimize hyperparameters. Data was split with the 70-20-10 predefined split provided in \textsc{OpenGraphXAI}. All models selected for testing achieved a minimum of a 0.92 F1 score on the validation set. Graph explainers were used to explain the predictions of the selected models on the test sets. Explainer hyperparameters have been set to the defaults provided in \cite{Fey/Lenssen/2019} when applicable. Additional details on model training, including computational resources used, are provided in Appendix \ref{appx:exp-details}. 

\subsection{Results}
\label{sec:results}

Table~\ref{tab:plausibility} presents the performance evaluation of the assessed explainers.
The results underscore the effectiveness of the CAM explainer, which achieves the highest performance in $13$ out of $15$ datasets, with exceptions being \texttt{charlie} and \texttt{golf}.
Specifically, Integrated Gradient demonstrates superior performance on the \texttt{charlie} dataset, while GNNExplainer excels on the \texttt{golf} dataset.

\begin{figure}
\centering
\begin{subfigure}{.33\textwidth}
  \centering
  \resizebox{!}{2.5cm}{\includegraphics[width=.99\textwidth]{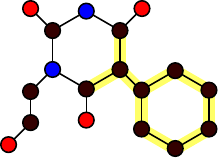}}
  \caption{Ground-truth}
  \label{fig:Ground truth mask}
\end{subfigure}%
\begin{subfigure}{.33\textwidth}
  \centering
  \resizebox{!}{2.5cm}{\includegraphics[width=.99\textwidth]{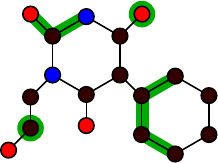}}
  \caption{Random explainer}
  \label{fig:RandomExplainer mask}
\end{subfigure}%
\begin{subfigure}{.33\textwidth}
  \centering
  \resizebox{!}{2.5cm}{\includegraphics[width=.99\textwidth]{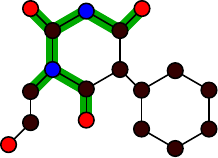}}
  \caption{Saliency}
  \label{fig:Saliency mask}
\end{subfigure}

\vspace{1em}

\begin{subfigure}{.33\textwidth}
  \centering
  \resizebox{!}{2.5cm}{\includegraphics[width=.99\textwidth]{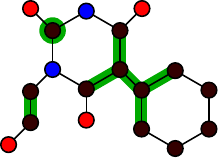}}
  \caption{Integrated Gradients}
  \label{fig:Integrated Gradients masks}
\end{subfigure}%
\begin{subfigure}{.33\textwidth}
  \centering
  \resizebox{!}{2.5cm}{\includegraphics[width=.99\textwidth]{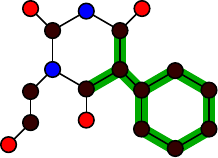}}
  \caption{CAM}
  \label{fig:Class Activation Mapping mask}
\end{subfigure}%
\begin{subfigure}{.33\textwidth}
  \centering
  \resizebox{!}{2.5cm}{\includegraphics[width=.99\textwidth]{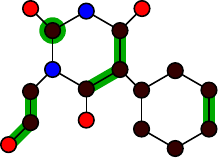}}
  \caption{GNNExplainer}
  \label{fig:GNNExplainer mask}
\end{subfigure}

\caption{Explanation masks computed by the tested explainers on a test graph from the \texttt{mike} dataset. Only the top 9 nodes (GT dimension) are highlighted for each mask.} \label{fig:ExplMasks}
\end{figure}

A qualitative assessment of the computed explanations can be conducted by examining Figure~\ref{fig:ExplMasks}, where the GT mask of a \texttt{juliett} dataset sample is compared to the explanations provided by all explainers.
In this instance, the CAM explanation mask exhibits a visually superior alignment compared to the others. The strong performance of the CAM method can be attributed to its model-specific nature, as its importance scores are directly derived from the activations of the model’s units. A second example is shown in supplemental Figure~\ref{fig:ExplMasks2}.

\begin{table}
	\caption{Plausibility results for a set of explainers on the \textsc{OpenGraphXAI} benchmarks with associated standard deviation. Best results are boldfaced.}
	\label{tab:plausibility}
	\centering
    \footnotesize
	\begin{tabular}{lcccccc}
		\toprule
		\textbf{Task} & \textbf{Class} & Random & Saliency & IntGrad & CAM & GNNExpl \\
		\midrule
		\multirow{2}{*}{\texttt{alfa}} & 0 & 0.524$\pm$0.131 & 0.135$\pm$0.175 & 0.921$\pm$0.088 & \textbf{0.942}$\pm$0.068 & 0.673$\pm$0.127\\
		& 1 & 0.481$\pm$0.108 & 0.406$\pm$0.185 & 0.616$\pm$0.113 & \textbf{0.642}$\pm$0.144 & 0.450$\pm$0.138  \\
        \midrule
		\multirow{2}{*}{\texttt{bravo}} & 0 & 0.516$\pm$0.129 & 0.361$\pm$0.253 & 0.597$\pm$0.137 & \textbf{0.697}$\pm$0.157 & 0.637$\pm$0.135 \\
		& 1 & 0.502$\pm$0.169 & 0.702$\pm$0.163 & 0.500$\pm$0.017 & \textbf{0.988}$\pm$0.038 & 0.767$\pm$0.226 \\
        \midrule
		\multirow{2}{*}{\texttt{charlie}} & 0 & 0.501$\pm$0.276 & 0.924$\pm$0.148 & \textbf{1.000}$\pm$0.000 & 0.883$\pm$0.110 & 0.643$\pm$0.351 \\
		& 1 & 0.489$\pm$0.112 & 0.330$\pm$0.144 & 0.406$\pm$0.060 & \textbf{0.599}$\pm$0.092 & 0.385$\pm$0.138 \\
        \midrule
		\multirow{2}{*}{\texttt{delta}} & 0 & 0.485$\pm$0.121 & 0.112$\pm$0.132 & 0.448$\pm$0.113 & \textbf{0.900}$\pm$0.113 & 0.494$\pm$0.146 \\
		& 1 & 0.492$\pm$0.119 & 0.223$\pm$0.212 & 0.460$\pm$0.089 & \textbf{0.898}$\pm$0.079 & 0.520$\pm$0.130 \\
        \midrule
		\multirow{2}{*}{\texttt{echo}} & 0 & 0.484$\pm$0.115 & 0.090$\pm$0.145 & 0.726$\pm$0.109 & \textbf{0.856}$\pm$0.151 & 0.670$\pm$0.139 \\
		& 1 & 0.495$\pm$0.117 & 0.584$\pm$0.184 & 0.526$\pm$0.124 & \textbf{0.706}$\pm$0.154 & 0.431$\pm$0.189 \\
        \midrule
		\texttt{foxtrot} & 1 & 0.491$\pm$0.123 & 0.181$\pm$0.130 & 0.743$\pm$0.099 & \textbf{0.930}$\pm$0.076 & 0.527$\pm$0.233 \\
        \midrule
		\texttt{golf} & 0 & 0.519$\pm$0.156 & 0.039$\pm$0.089 & 0.681$\pm$0.148 & 0.683$\pm$0.167 & \textbf{0.874}$\pm$0.180 \\
        \midrule
		\texttt{hotel} & 0 & 0.431$\pm$0.199 & 0.190$\pm$0.203 & 0.482$\pm$0.201 & \textbf{0.871}$\pm$0.247 & 0.486$\pm$0.187 \\
        \midrule
		\texttt{india} & 0 & 0.480$\pm$0.159 & 0.245$\pm$0.240 & 0.633$\pm$0.198 & \textbf{0.918}$\pm$0.102 & 0.509$\pm$0.190 \\
        \midrule
		\texttt{juliett} & 0 & 0.464$\pm$0.212 & 0.494$\pm$0.309 & 0.606$\pm$0.244 & \textbf{0.844}$\pm$0.226 & 0.540$\pm$0.192 \\
        \midrule
		\texttt{kilo} & 0 & 0.510$\pm$0.151 & 0.796$\pm$0.161 & 0.833$\pm$0.073 & \textbf{0.947}$\pm$0.098 & 0.570$\pm$0.238 \\
        \midrule
		\texttt{lima} & 0 & 0.502$\pm$0.136 & 0.123$\pm$0.097 & 0.764$\pm$0.088	 & \textbf{0.955}$\pm$0.051 & 0.575$\pm$0.254 \\
        \midrule
		\texttt{mike} & 0 & 0.497$\pm$0.154 & 0.138$\pm$0.182 & 0.678$\pm$0.167 & \textbf{0.840}$\pm$0.184 & 0.332$\pm$0.178 \\
        \midrule
		\texttt{november} & 0 & 0.531$\pm$0.173 & 0.018$\pm$0.041 & 	0.603$\pm$0.049 & \textbf{1.000}$\pm$0.004 & 0.408$\pm$0.170 \\
        \midrule
		\texttt{oscar} & 0 & 0.490$\pm$0.137 & 0.183$\pm$0.207 & 0.938$\pm$0.082 & \textbf{0.973}$\pm$0.053 & 0.501$\pm$0.180\\
		\bottomrule
	\end{tabular}
\end{table}

\subsubsection{Statistical significance}

\begin{figure}[t!]
    \centering
    \includegraphics[width=.85\textwidth]{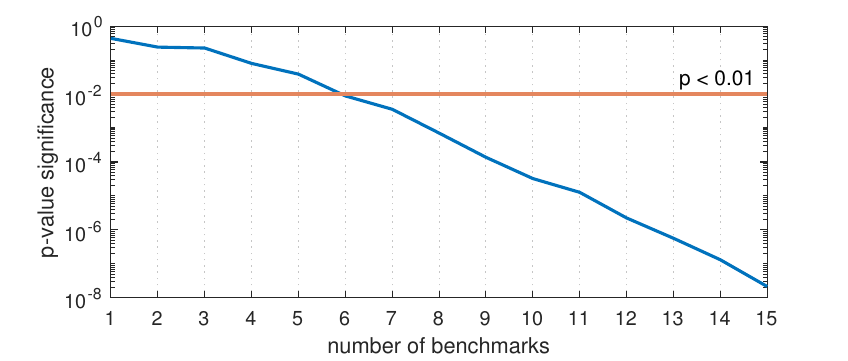}
    \caption{Statistical significance of the explainer ranking as $p$-value of the Friedman test \citep{Demsar2006}, varying the number of benchmarks considered. At least $7$ benchmarks are required for $p < 0.01$ significant results. \textsc{OpenGraphXAI} benchmarks allow $p \approx 10^{-8}$, or $>5\sigma$ significance.}
    \label{fig:friedman}
\end{figure}

To test the significance of the ranking of the explainer methods in our experiments, we apply the Friedman test \citep{Demsar2006}.
The null hypothesis that the XAI methods' ranking is random is rejected with $p < 10^{-7}$, or a confidence level $> 5\sigma$, in other terms.
In Figure~\ref{fig:friedman} we report the $p$-value that the Friedman test would have returned if less than the full 15 \textsc{OpenGraphXAI} had been considered in the experimental evaluation.
As at least $7$ benchmarks are required for $p < 0.01$ significant results, our contribution of a large collection of benchmarks is fundamental to enable researchers to drive a meaningful experimental evaluation of graph-XAI methods, a task hitherto unachievable due to the benchmark scarcity.

\begin{figure}[t!]
    \centering
    \includegraphics[width=.75\textwidth]{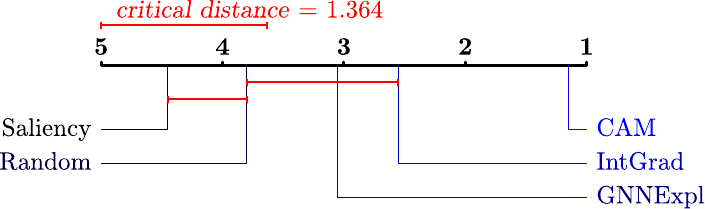}
    \caption{CAM ranks as the best explainer with significance level $\alpha = 0.05$ according to the \textit{post-hoc} Nemenyi test \citep{Demsar2006}. Statistically compatible average rankings are joined in cliques.}
    \label{fig:explainer-ranking}
\end{figure}

In Figure~\ref{fig:explainer-ranking} we report the critical difference diagram for the \textit{post-hoc} Nemenyi test with significance level $\alpha = 0.05$.
This test allows us to compare the ranking of graph explainers in terms of plausibility: the performance of two explainers is significantly different if the corresponding average ranks differ by at least the critical difference \citep{Demsar2006}.
The top-ranked explainer method is CAM, which is also the statistically significant best explainer.
IntGrad and GNNExpl rank below CAM beyond the critical distance, and together with Saliency, are compatible with the Random explainer.

\section{Conclusions}
\label{sec:conclusion}

We have introduced a method, based on the WL coloring algorithm, to systematically generate graph-XAI benchmarks from existing graph classification datasets.
We applied this methodology to build a new benchmark collection, called \textsc{OpenGraphXAI}, that consists of 15 graph-XAI tasks derived from 11 real-world molecular datasets, which can be extended to more than 2000 benchmark datasets with annotated GTs.
This collection provides a suite of XAI tasks to challenge the ability of graph explainers to identify the GTs based on structural patterns of different complexity that appear in real-world graphs.
Our contribution strikes a balance between the chronic scarcity of graph-XAI datasets with GTs that are annotated by domain experts, and the simplistic nature of current synthetic datasets, which are based on random graphs augmented with trivial, non-realistic motifs.
To illustrate a use case, we perform an experimental evaluation of widely adopted XAI methods for graph classification.
The statistical significance analysis makes it evident that a large collection of benchmarks is necessary to draw rigorous insights, with the ultimate goal of fostering continuous progress in graph-XAI research.

\backmatter

\section*{Declarations}

\bmhead{Funding} 
Research partly supported by: PNRR, PE00000013, “FAIR - Future Artificial Intelligence Research”, Spoke 1, funded by European Commission under NextGeneration EU programme (CUP: B53D23026250001); Project DEEP-GRAPH, funded by the Italian Ministry of University and Research (MUR) PRIN 2022 (project code: 2022YLRBTT, CUP: I53C24002440006); Project PAN-HUB, funded by the Italian Ministry of Health (POS 2014–2020, project ID: T4-AN-07, CUP: I53C22001300001).

\bmhead{Conflicts of interest} 
The authors have no relevant financial or non-financial interests to disclose.

\bmhead{Ethics approval and consent to participate} 
Not applicable.

\bmhead{Consent for publication} 
Not applicable.

\bmhead{Data availability} 
We release the \textsc{OpenGraphXAI} benchmark suite in plain JSON format at the following URL:  \url{https://www.kaggle.com/datasets/dtortorella/ogx-benchmarks}.

\bmhead{Code availability}
Code is hosted at the following URL:  \url{https://github.com/OpenGraphXAI/benchmarks}

\bmhead{Author contributions} 
AM conceived the study. MF and DT wrote the code and performed the experiments. MF, DT, and MP wrote the initial draft of the manuscript. All authors performed the analysis, read, and approved the final draft.

\noindent

\begin{appendices}

\section{Ground truths}
\label{appx:ogx-gts}

\begin{figure}[p] 
\centering
\begin{subfigure}[b]{.33\textwidth}
  \centering
  \resizebox{.5\textwidth}{!}{\includegraphics[width=.99\textwidth]{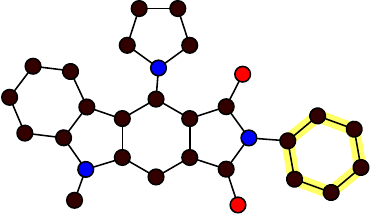}}
  \caption{\texttt{alfa} (class 0)}
  \label{fig:alfa0}
\end{subfigure}%
\begin{subfigure}[b]{.33\textwidth}
  \centering
  \resizebox{.6\textwidth}{!}{\includegraphics[width=.99\textwidth]{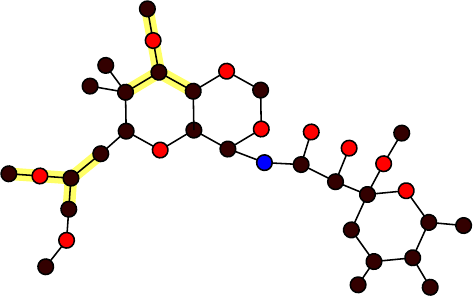}}
  \caption{\texttt{alfa} (class 1)}
  \label{fig:alfa1}
\end{subfigure}%
\begin{subfigure}[b]{.33\textwidth}
  \centering
  \resizebox{.5\textwidth}{!}{\includegraphics[width=.99\textwidth]{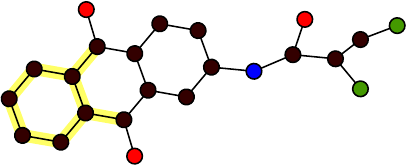}}
  \caption{\texttt{bravo} (class 0)}
  \label{fig:bravo0}
\end{subfigure}

\begin{subfigure}[b]{.33\textwidth}
  \centering
    \resizebox{.6\textwidth}{!}{\includegraphics[width=.99\textwidth]{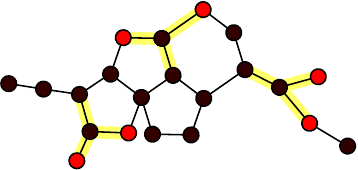}}
  \caption{\texttt{bravo} (class 1)}
  \label{fig:bravo1}
\end{subfigure}%
\begin{subfigure}[b]{.33\textwidth}
  \centering
  \resizebox{.6\textwidth}{!}{\includegraphics[width=.99\textwidth]{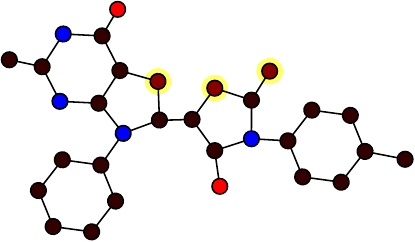}}
  \caption{\texttt{charlie} (class 0)}
  \label{fig:charlie0}
\end{subfigure}%
\begin{subfigure}[b]{.33\textwidth}
  \centering
  \resizebox{.7\textwidth}{!}{\includegraphics[width=.99\textwidth]{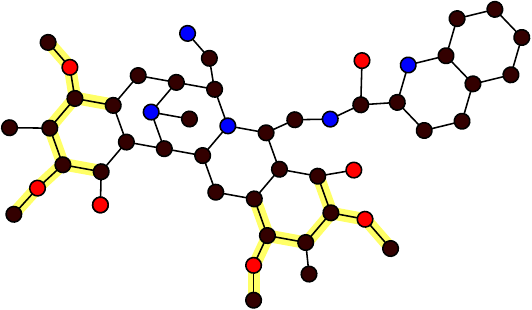}}
  \caption{\texttt{charlie} (class 1)}
  \label{fig:charlie1}
\end{subfigure}

\begin{subfigure}[b]{.33\textwidth}
  \centering
  \resizebox{.6\textwidth}{!}{\includegraphics[width=.99\textwidth]{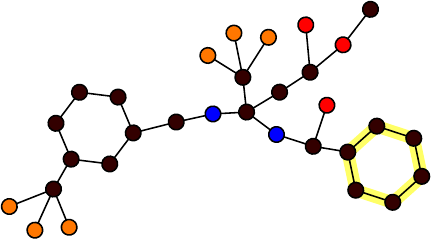}}
  \caption{\texttt{delta} (class 0)}
  \label{fig:delta0}
\end{subfigure}%
\begin{subfigure}[b]{.33\textwidth}
  \centering
  \resizebox{.5\textwidth}{!}{\includegraphics[width=.99\textwidth]{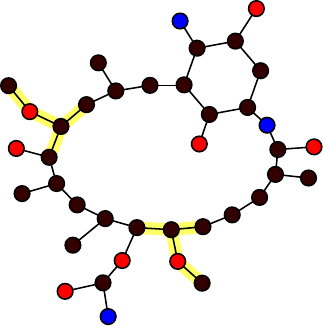}}
  \caption{\texttt{delta} (class 1)}
  \label{fig:delta1}
\end{subfigure}%
\begin{subfigure}[b]{.33\textwidth}
  \centering
  \resizebox{.55\textwidth}{!}{\includegraphics[width=.99\textwidth]{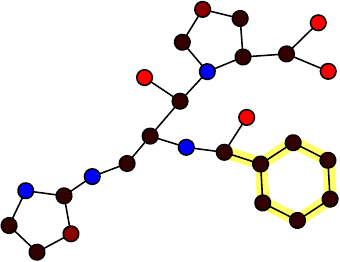}}
  \caption{\texttt{echo} (class 0)}
  \label{fig:echo0}
\end{subfigure}

\begin{subfigure}[b]{.33\textwidth}
  \centering
  \resizebox{.5\textwidth}{!}{\includegraphics[width=.99\textwidth]{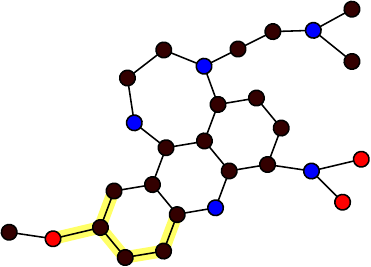}}
  \caption{\texttt{echo} (class 1)}
  \label{fig:echo1}
\end{subfigure}%
\begin{subfigure}[b]{.33\textwidth}
  \centering
  \resizebox{.6\textwidth}{!}{\includegraphics[width=.99\textwidth]{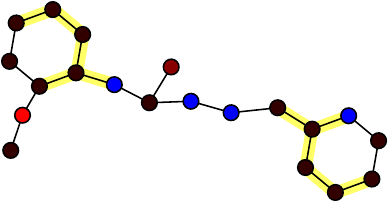}}
  \caption{\texttt{foxtrot} (class 1)}
  \label{fig:foxtrot1}
\end{subfigure}%
\begin{subfigure}[b]{.33\textwidth}
  \centering
  \resizebox{.5\textwidth}{!}{\includegraphics[width=.99\textwidth]{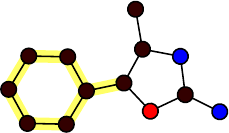}}
  \caption{\texttt{golf} (class 0)}
  \label{fig:golf0}
\end{subfigure}

\begin{subfigure}[b]{.33\textwidth}
  \centering
  \resizebox{.55\textwidth}{!}{\includegraphics[width=.99\textwidth]{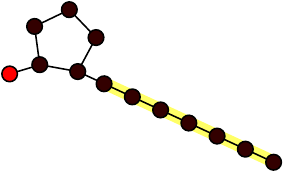}}
  \caption{\texttt{hotel} (class 0)}
  \label{fig:hotel0}
\end{subfigure}%
\begin{subfigure}[b]{.33\textwidth}
  \centering
  \resizebox{.9\textwidth}{!}{\includegraphics[width=.99\textwidth]{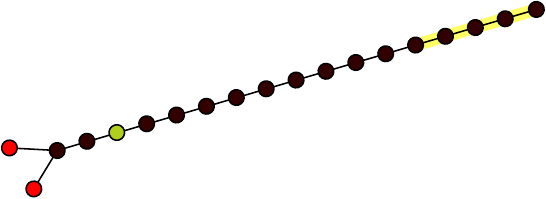}}
  \caption{\texttt{india} (class 0)}
  \label{fig:india0}
\end{subfigure}%
\begin{subfigure}[b]{.33\textwidth}
  \centering
  \resizebox{.9\textwidth}{!}{\includegraphics[width=.99\textwidth]{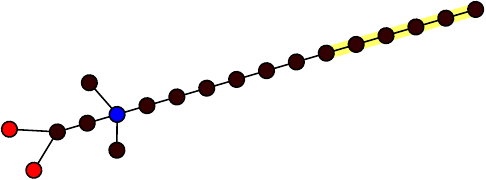}}
  \caption{\texttt{juliett} (class 0)}
  \label{fig:juliett0}
\end{subfigure}

\begin{subfigure}[b]{.33\textwidth}
  \centering
  \resizebox{.5\textwidth}{!}{\includegraphics[width=.99\textwidth]{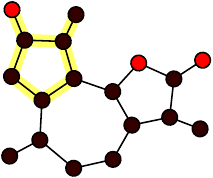}}
  \caption{\texttt{kilo} (class 0)}
  \label{fig:kilo0}
\end{subfigure}%
\begin{subfigure}[b]{.33\textwidth}
  \centering
  \resizebox{.6\textwidth}{!}{\includegraphics[width=.99\textwidth]{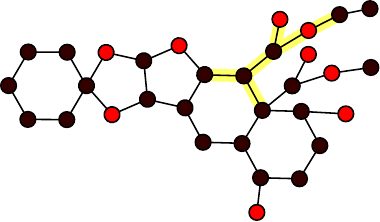}}
  \caption{\texttt{lima} (class 0)}
  \label{fig:lima0}
\end{subfigure}%
\begin{subfigure}[b]{.33\textwidth}
  \centering
  \resizebox{.6\textwidth}{!}{\includegraphics[width=.99\textwidth]{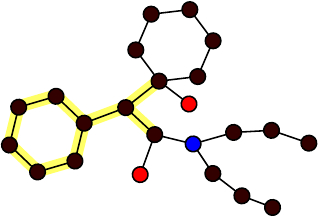}}
  \caption{\texttt{mike} (class 0)}
  \label{fig:mike0}
\end{subfigure}

\begin{subfigure}[b]{.5\textwidth}
  \centering
  \resizebox{.6\textwidth}{!}{\includegraphics[width=.99\textwidth]{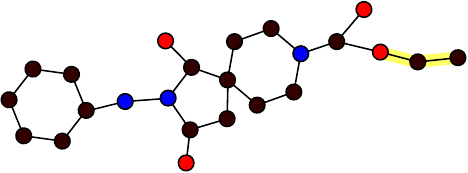}}
  \caption{\texttt{november} (class 0)}
  \label{fig:november0}
\end{subfigure}%
\begin{subfigure}[b]{.5\textwidth}
  \centering
  \resizebox{.3\textwidth}{!}{\includegraphics[width=.99\textwidth]{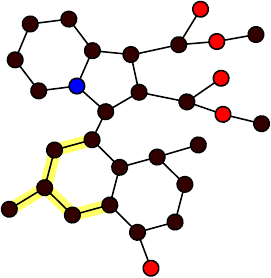}}
  \caption{\texttt{oscar} (class 0)}
  \label{fig:oscar0}
\end{subfigure}
\caption{Examples from the datasets in \textsc{OpenGraphXAI}. GT explanations are highlighted in yellow.}
\label{fig:ogx-gt2}
\end{figure}

In Figure \ref{fig:ogx-gt2}, we show one example for each class available in the \textsc{OpenGraphXAI} benchmark, across the different tasks.

\section{Hyperparameters}
\label{appx:exp-details}

During the experiments, The learning rate was tuned from $\{0.001, 0.0001\}$, the number of message passing iterations (corresponding to the number of architectural layers) from \{1,2,3,4,5\}, the number of units for each layer from $\{32, 64\}$ and the weight decay factor from $\{0.001, 0.0001\}$. We used the Adam optimizer \citep{kingma2014adam}, training for a maximum of 1500 epochs with an early stopping patience of 30 to prevent overfitting. Table~\ref{tab:selection} shows the scores obtained by the selected models on the test set, and their respective hyperparameters. All experiments have been conducted on a machine with 2 Xeon Gold 6238R of 28 cores and a base clock of 2.20 GHz each, 256 GB of RAM, and 2 GPU NVIDIA A30 featuring 3584 CUDA cores.

\begin{table}[h!]
\caption{Hyperparameters and performance of the selected GNN models across the different \textsc{OpenGraphXAI} tasks.}\label{tab:selection}
\centering
\begin{tabular}{lccccccc}
\toprule
Task     & LR & \#L & ED & WD & Training F1 & Validation F1 & Test F1 \\ \midrule
\texttt{alfa}     & $1 \times 10^{-4}$        & 3                & 64                   & $1 \times 10^{-4}$       & 0.969    & 0.975         & 0.983   \\
\texttt{bravo}    & $1 \times 10^{-3}$         & 4                & 64                   & $1 \times 10^{-4}$       & 0.998    & 0.991         & 0.991   \\
\texttt{charlie}  & $1 \times 10^{-3}$         & 1                & 32                   & $1 \times 10^{-4}$       & 1.000    & 1.000         & 1.000   \\
\texttt{delta}    & $1 \times 10^{-3}$         & 4                & 32                   & $1 \times 10^{-3}$        & 1.000    & 1.000         & 0.973   \\
\texttt{echo}     & $1 \times 10^{-3}$         & 4                & 32                   & $1 \times 10^{-4}$       & 0.958    & 0.952         & 0.919   \\
\texttt{foxtrot}  & $1 \times 10^{-3}$         & 5                & 64                   & $1 \times 10^{-3}$        & 0.982    & 0.922         & 0.936   \\
\texttt{golf}     & $1 \times 10^{-3}$         & 5                & 64                   & $1 \times 10^{-4}$       & 0.974    & 0.927         & 0.926   \\
\texttt{hotel}    & $1 \times 10^{-3}$         & 5                & 64                   & $1 \times 10^{-4}$       & 0.995    & 1.000         & 0.989   \\
\texttt{india}    & $1 \times 10^{-3}$         & 4                & 32                   & $1 \times 10^{-3}$        & 0.992    & 0.975         & 0.983   \\
\texttt{juliett}  & $1 \times 10^{-3}$         & 5                & 32                   & $1 \times 10^{-3}$        & 0.966    & 0.989         & 0.989   \\
\texttt{kilo}     & $1 \times 10^{-3}$         & 4                & 32                   & $1 \times 10^{-3}$        & 0.999    & 0.995         & 1.000   \\
\texttt{lima}     & $1 \times 10^{-3}$         & 4                & 64                   & $1 \times 10^{-3}$        & 0.996    & 0.994         & 0.998   \\
\texttt{mike}     & $1 \times 10^{-3}$         & 5                & 64                   & $1 \times 10^{-3}$        & 0.982    & 0.959         & 0.955   \\
\texttt{november} & $1 \times 10^{-3}$         & 4                & 64                   & $1 \times 10^{-3}$        & 0.999    & 1.000         & 1.000   \\
\texttt{oscar}    & $1 \times 10^{-3}$         & 5                & 32                   & $1 \times 10^{-4}$       & 0.999    & 0.999         & 0.998   \\ \bottomrule
\end{tabular}
{\footnotesize LR: learning rate, \#L: number of layers, ED: embedding dimension, WD: weight decay.}
\end{table}

\section{Additional explanation example}
\label{appx:exp-additional}

\begin{figure}[h!] 
\centering
\begin{subfigure}{.5\textwidth}
  \centering
  \resizebox{5cm}{!}{\includegraphics[width=.99\textwidth]{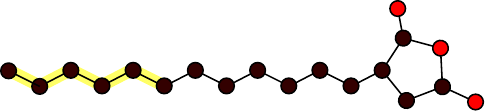}}
  \caption{Ground-truth}
  \label{fig:Ground truth mask}
\end{subfigure}%
\begin{subfigure}{.5\textwidth}
  \centering
  \resizebox{5cm}{!}{\includegraphics[width=.99\textwidth]{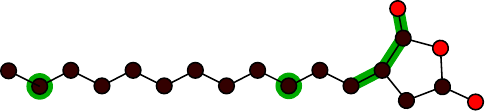}}
  \caption{Random explainer}
  \label{fig:RandomExplainer mask}
\end{subfigure}

\begin{subfigure}{.5\textwidth}
  \centering
  \resizebox{5cm}{!}{\includegraphics[width=.99\textwidth]{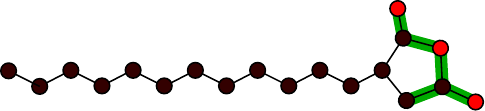}}
  \caption{Saliency}
  \label{fig:Saliency mask}
\end{subfigure}%
\begin{subfigure}{.5\textwidth}
  \centering
  \resizebox{5cm}{!}{\includegraphics[width=.99\textwidth]{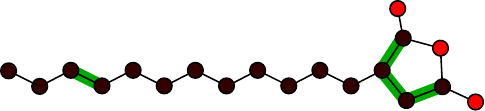}}
  \caption{Integrated Gradients}
  \label{fig:Integrated Gradients masks}
\end{subfigure}

\begin{subfigure}{.5\textwidth}
  \centering
  \resizebox{5cm}{!}{\includegraphics[width=.99\textwidth]{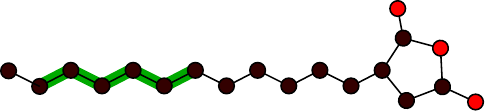}}
  \caption{Class Activation Mapping (CAM)}
  \label{fig:Class Activation Mapping mask}
\end{subfigure}%
\begin{subfigure}{.5\textwidth}
  \centering
  \resizebox{5cm}{!}{\includegraphics[width=.99\textwidth]{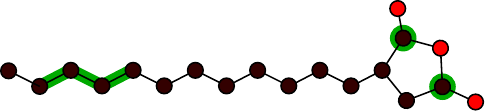}}
  \caption{GNNExplainer}
  \label{fig:GNNExplainer mask}
\end{subfigure}

\caption{Explanation masks computed by the tested explainers on a test graph of the \texttt{juliett} dataset. Only the top 6 nodes (GT dimension) are highlighted for each mask.} \label{fig:ExplMasks2}
\end{figure}

In Figure~\ref{fig:ExplMasks2}, we present an additional example of explanations computed by the explainers evaluated in our experiments.

\end{appendices}


\clearpage

\bibliography{sn-bibliography}

\end{document}